\documentclass[11pt,a4paper]{article}
\usepackage{authblk}
\usepackage[hyperref]{acl2019}
\usepackage{times}
\usepackage{latexsym}
\usepackage{float}
\usepackage{url}
\usepackage{array}
\usepackage{booktabs}
\usepackage{arydshln}
\usepackage{graphicx}
\usepackage{amssymb}
\usepackage{amsmath}
\usepackage{bm}
\usepackage{multirow}
\usepackage[bb=boondox]{mathalfa}
\usepackage{xspace}
\usepackage{color,soul}

\newcolumntype{L}[1]{>{\raggedright\let\newline\\\arraybackslash\hspace{0pt}}m{#1}}
\newcolumntype{C}[1]{>{\centering\let\newline\\\arraybackslash\hspace{0pt}}m{#1}}
\newcolumntype{R}[1]{>{\raggedleft\let\newline\\\arraybackslash\hspace{0pt}}m{#1}}

\newcommand{\bsy}[1]{\boldsymbol{#1}}
\newcommand{\mb}[1]{\mathbf{#1}}
\newcommand{\cev}[1]{\reflectbox{\ensuremath{\vec{\reflectbox{\ensuremath{#1}}}}}}
\DeclareMathOperator*{\argmax}{arg\,max}

\DeclareMathOperator*{\AGGREGATE}{Aggregate}
\DeclareMathOperator*{\softmax}{Softmax}
\newcommand{\SAT}{SA-T\xspace}

\aclfinalcopy

\def\aclpaperid{556} 

\title{Extracting Symptoms and their Status from Clinical Conversations}

\author{Nan Du\thanks{\ All the authors contributed equally.}}
\author{Kai Chen}
\author{Anjuli Kannan}
\author{Linh Tran}
\author{Yuhui Chen}
\author{Izhak Shafran}
\affil{\{dunan, kaichen, anjuli, tranlm, yuhuic, izhak\}@google.com}
\affil{Google Inc.}

\date{}

\begin{document}
\maketitle

\begin{abstract}
This paper describes novel models tailored for a new application, that of extracting the symptoms mentioned in clinical conversations along with their status. Lack of any publicly available corpus in this privacy-sensitive domain led us to develop our own corpus, consisting of about 3K conversations annotated by professional medical scribes. We propose two novel deep learning approaches to infer the symptom names and their status: (1) a new hierarchical span-attribute tagging (\SAT) model, trained using curriculum learning, and (2) a variant of sequence-to-sequence model which decodes the symptoms and their status from a few speaker turns within a sliding window over the conversation. This task stems from a realistic application of assisting medical providers in capturing symptoms mentioned by patients from their clinical conversations. To reflect this application, we define multiple metrics. From inter-rater agreement, we find that the task is inherently difficult. We conduct comprehensive evaluations on several contrasting conditions and observe that the performance of the models range from an F-score of 0.5 to 0.8 depending on the condition. Our analysis not only reveals the inherent challenges of the task, but also provides useful directions to improve the models.
\end{abstract}

\section{Introduction}
\label{sec:intro}
In recent years, hospitals and clinics across the United States have been coaxed and cajoled into adopting Electronic Health Records through public policies and insurance requirements. This has led to the unforeseen side-effect of placing a disproportionate burden of documentation on physicians, causing burnouts among them~\cite{HBR2018,Atlantic2018}. One study found that full-time primary care physicians spent about 4.5 hours of an 11-hour workday interacting with the clinical documentation systems, and yet were still unable to finish their documentations and had to spend an additional 1.4 hours after normal clinical hours \cite{Arndt:2017}.

Speech and natural language processing are now sufficiently mature that there has been considerable interest, both in academia and industry, to investigate how these technologies can be exploited to simplify the task of documentation, and to allow physicians to dedicate more time to patients. While domain-specific ASR systems that allow doctors to dictate notes have been around for a while, recent work~\cite{PatDavPan18, FinleyEdwards:2018, FinSalSad18} has begun to address more challenging tasks, such as extracting relevant information directly from doctor-patient conversations.

In this work, we investigated the task of inferring symptoms mentioned in clinical conversations, along with whether patients have experienced them or not. Our contributions include: (i) defining the task, including the annotation scheme for labeling the clinical conversations and the evaluation metrics to measure model performance (Section~\ref{sec:task}); (ii) two novel deep learning models to solve this task (Section~\ref{sec:models}); (iii) comprehensive empirical evaluations in different contrasting conditions (Section~\ref{sec:expts}), and (iv) analysis of the performance of the models that provides meaningful insights for further improvements (Section~\ref{sec:analysis}). 

\section{Related Work}
\label{sec:related}

On the topic of information extraction from medical text, one of the earliest public-domain task is the {\it i2b2 challenge}, defined on a small corpus of written discharge summaries that consists of 394 reports for training, 477 for test, and 877 for evaluation \cite{Uzuner:2011}. Given the small amount of training data, not surprisingly, a disproportionately large number of teams fielded rule-based systems. CRF-based systems however did better even with the limited amount of training data. Being a written domain task, they benefited from section headings and other cues that are unavailable in doctor-patient conversations. 
For a wider survey of extracting clinical information from written clinical documents, see~\cite{Liu:2012}.

There are very few publications on processing clinical conversations. One noteworthy recent work extracts entities using a multi-stage approach~\cite{FinleyEdwards:2018}. They use two-level hierarchical model, modeling word sequences and sentence sequences, to classify sentences into the sections in a clinical note they belong to. The extracted sentences are then processed using a variety of heuristics such as partial string matching with an ontology, regular expressions, and other task-specific heuristics. One would imagine sentences taken out of context of a dialog are prone to misinterpretation and they do not elaborate on how that is overcome. Moreover, their system cannot be optimized end-to-end. 

Other related work includes normalizing the terms and mapping them to external databases such as Unified Medical Language System (UMLS) and specific sub-tasks such as negation detection, which are outside the scope of this work~\cite{HappePBCB03,Névéol14thequaero,Lowe2007}. 

\section{The Symptom Extraction Task}
\label{sec:task}
We begin the description of our task by introducing the corpus, the annotation paradigm, and the evaluation metrics. 
\subsection{Corpus Description}
\label{ssec:corpus}
Our unlabeled corpus consists of 90k de-identified and manually transcribed audio recordings of clinical conversations between physicians and patients, typically about 10 minutes long. A few of the conversations also contain speech from nurses, caregivers, spouses and other attendees.

The annotation guidelines were developed by a team of professional medical scribes, physicians and natural language processing experts. Two primary categories of labels were annotated: the symptoms being discussed and their status. An ontology of 186 symptoms were defined (e.g., vomiting, nausea, diarrhea), each belonging to one of 14 body systems (e.g., gastrointestinal, musculo-skeletal, cardiovascular). For each symptom, annotators were instructed to associate a status that denotes whether the patient has experienced it or not. An additional catch-all category was defined to include symptoms whose status cannot be conclusively inferred from the conversation or which are not relevant to the clinical note. Thus, status may have one of the three values: {\it experienced}, {\it not experienced}, and {\it other}. In an utterance, ``{\it I have a \underline{back pain}}'', the underlined phrase will be assigned the tuple: ({\it sym:musculo-skeletal:pain, experienced}). The top three symptoms in the corpus are: musculo-skeletal pain, shortness of breath and cough.

Of the 90K encounters, we chose to focus on primary care visits. A team of 18 professional scribes was trained on the guidelines. They labeled the manual transcripts of 2,950 conversations, which were partitioned into training (1,950), development (500) and test (500) sets. The entire labeled corpus contains 5M tokens in 615K sentences with 92K labels.

To account for variation across scribes, we randomly assigned 3 scribes to label each of the conversations in the development (500) and test (500) sets. The inter-labeler agreement in terms of Cohen's kappa is 0.4 on the development set. Further analyses showed that the low score was largely due to (i) the ambiguous and informal ways that patients and doctors discuss symptoms, (ii) that human scribes often disagree on which one of closely related labels to pick, and (iii) that human scribes often disagree on the span of text to label.

\subsection{Evaluation Metrics}
\label{ssec:metrics}
In clinical conversations, the symptoms may be mentioned multiple times, paraphrased differently, but still may appear in the clinical notes only once. So, we chose to evaluate them at the conversation levels using two metrics.

\noindent
{\bf Unweighted metric}: In this metric, we account only for the unique symptoms and ignore the number of times they were mentioned. The set of events in the inferred output was compared against the set in the reference to compute the precision and recall for each conversation before averaging across all conversations.

\noindent
{\bf Weighted metric}: The symptoms that are mentioned more often in a conversation are likely to be more important. In this metric, each symptom is weighted by its frequency: precision is weighted by the frequency of the predictions, while recall is weighted by the frequency of the reference.

\section{Models}
\label{sec:models}
We developed two novel neural network model architectures for this task: 1) a span-attribute model that is similar in spirit to a tagging model but works well on our large label space, and 2) a sequence-to-sequence (Seq2Seq) model~\cite{Sutskever14, ChoMerrienboerGulcehreEtAl14} that is designed to infer symptoms that are described informally across a few conversation turns.

\subsection{Span-Attribute Tagging (\SAT) Model}
\label{ssec:tagging}
A common solution for this task is a tagging model, where the word sequences are represented by word and/or character embeddings and fed into a sequence of layers consisting of a bidirectional layer, a softmax layer and a conditional random field (CRF) to predict the BIO-style tags~\cite{ColWesBot11, HuaXuYu15, MaHov16, ChiNic16, LamBalSub16, PetAmmBha17, YanSalCoh17, ChaHuSha18}. However, in our task, the tags need to identify not only the symptom names associated with the words but also the status. This can be accomplished in a tagging model using a label space that is the Cartesian product of both the symptom names and their status. Unfortunately, this Cartesian space turns out to quite large in our task (186 x 3). Tagging models perform well when the set of tags is reasonably small (e.g., named entity recognition and part of speech tagging), but not so well when the set of tags is large. Moreover, in our case, given the limited corpus size, modeling the cross-product space leads to data sparsity. 

\begin{figure}[t]
\centering
\includegraphics[width=\linewidth]{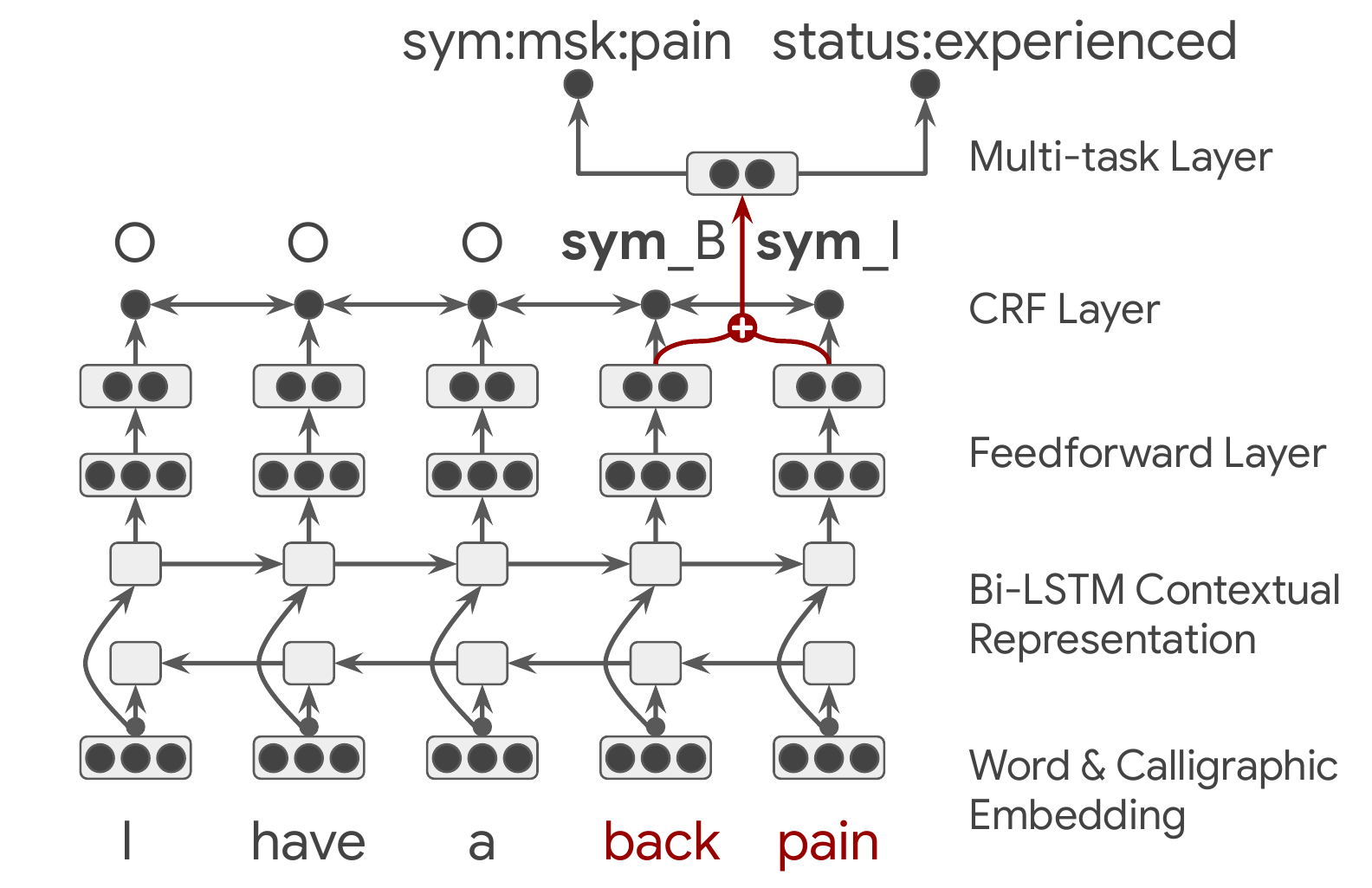} \caption{The architecture of Span-Attribute Tagging (\SAT) Model, illustrating the span extraction layer followed by the attribute tagging layer.}
\label{figure:tagging-architecture}
\end{figure}

For tackling this challenge of data sparsity, we reformulate the problem from a novel hierarchical perspective. Unlike the conventional tagging model, where at each input token the model has to pick the best candidate label from the full label space, we break this into two stages. We first identify the span of interest using a generic tag set with a very small label set of just three elements, \{{\it sym\_B, sym\_I, O}\}. This simplifies the computational cost of inferring over sequence, which allows us to employ the CRF layer. 
Moreover, it alleviates the data sparsity problem by pooling all the labels to identify all spans of interest. In the second stage, we predict the attributes associated with the span using contextual features of arbitrary complexity without encumbering the inference over the entire sequence. In addition, since our label space can be partitioned easily, we use two separate predictors, one for symptom name and one for status. These two stages are trained jointly in an end-to-end fashion using multi-task learning, as described later.

Figure~\ref{figure:tagging-architecture} illustrates this hierarchical perspective for our task. The first stage, which is akin to a conventional tagging model, identifies the span of interest -- {\it back pain} -- at the output of the CRF layer. The second stage utilizes the latent representation from the span and employs two separate predictors to classify the symptom name as {\it sym:msk:pain} and the status as {\it experienced}. In principle, these predictors can be more complex than a simple soft-max that we have used. We refer to this architecture as the Span-Attribute Tagging (\SAT) model. The two stages of the model are described in more details below. 

\paragraph{Span Extraction Layer} As mentioned before, this layer employs a conventional tagging model whose output is constrained to be just three elements of $\cal E =\{{\it sym\_B, sym\_I, O}\}$. The model is briefly described as follows. 

Let $\mathbf{x}$ be the embedding vector sequence corresponding to the input word sequence. From this sequence, we compute a sequence of latent contextual representations using a bidirectional LSTM, $\mathbf{h'}=[\vec{\mathbf{h}}(\mathbf{x}|\vec{\Theta}_{LSTM}),\cev{\mathbf{h}}(\mathbf{x}|\cev{\Theta}_{LSTM})]$. This latent contextual sequence is fed into a two-layer fully connected feed-forward network to obtain a final sequence of latent representation $\mathbf{h''} = MLP(\mathbf{h'}|\Theta_{FF})$. Given this feature representation $\mathbf{h}=(\bsy{h^{''}_1},\cdots,\bsy{h^{''}_N})$ and the target tag sequence $\mathbf{y}^e=(y_i \, | \, i=1, \dots, N; \, y_i \in \cal E$), the parameters of the model are learned by minimizing the negative log-likelihood $-\log P(\mb{y}^e|\mb{h})$. This is computed in terms of a compatibility function defined over any label sequence $\mb{y}$ and $\mb{h}$.
\begin{equation}
S(\mb{y}, \mb{h}) = \sum_{i=0}^N\mb{A}_{y_i, y_{i+1}} + \sum_{i=0}^N P(\mb{h}_i, \mb{y}_i)
\end{equation}

Of the two components, the first one estimates the probability of the label sequence in terms of the sum of first order Markov transition of the label sequence $\mb{y}$, computed from a learned transition matrix $A$ whose dimensions are ${\cal |E| \times |E|}$. The second component estimates the joint probability of the latent vector $\mb{h}_{i}$ and the corresponding label embedding $\mb{y}_{i}$, specifically, in terms of similarity measure $\mb{h}_{i}^\top \mb{y}_{i}$. 

Using the compatibility function, the loss for the task of recognizing the spans is estimated as $- S(\mathbf{y}^e, \mathbf{h}) + \log{\sum_{\mathbf{y}^\prime}\exp{(S(\mathbf{y}^\prime, \mathbf{h}))}}$, where $\mathbf{y}^\prime$ is any other possible sequence of labels. During training, $\log P(\mb{y}^e|\mb{h})$ is estimated using forward-backward algorithm, and during inference, the most probable sequence $\mb{y}^*=\argmax_{\mb{y}^\prime} P(\mb{y}^\prime|\mb{h})$ is computed using the Viterbi algorithm. 

\paragraph{Attribute Tagging Layer} Given the span, as mentioned before, we can potentially use a richer representation of the context to predict attributes than otherwise possible. A contextual representation is computed from the starting index $i$ and ending index $j$ of each span using a pooling function $\AGGREGATE(\cdot)$.
\begin{align}
    \bsy{h}^s_{ij} = \AGGREGATE(\bsy{h}_k|\bsy{h}_k\in\mb{h}, i\leq k < j) \label{eqn:context}
\end{align}
The pooling function can be implemented as simple as mean or sum, or as the hidden state of another encoder like BiLSTM, CNN or self-attention~\cite{transformer}. Given the span representation $\bsy{h}^s_{ij}$, we model the joint distribution of the symptom name and status as $P(y^{s_x}, y^{s_t}|\bsy{h}^s_{ij}) = P(y^{s_x}|\bsy{h}^s_{ij})P(y^{s_t}|\bsy{h}^s_{ij})$ with the assumption that they are independent. Then, the distribution over the symptom name for each span is a multinomial distribution $P(y^{s_x}=k|\bsy{h}^s_{ij}) = \softmax(\bsy{h}^s_{ij}|\Theta^{s_x})_k$.
Similarly, we can formulate the distribution over the symptom status as $P(y^{s_t}=m|\bsy{h}^s_{ij}) = \softmax(\bsy{h}^s_{ij}|\Theta^{s_t})_m$. Both $\Theta^{s_x}$ and $\Theta^{s_t}$ are model parameters. Finally, we can train the model end-to-end by minimizing the following loss function for each conversation:
\begin{align}
&\ell(\mb{y}^e,\{(y^{s_x},y^{s_t})\}|\mb{h}) = -\alpha\log P(\mb{y}^e|\mb{h})+\nonumber\\
&\sum_{\{(y^{s_x},y^{s_t})\}}-\log P(y^{s_x}|\mb{h})
- \log P(y^{s_t}|\mb{h}),
\end{align}
where $\{(y^{s_x},y^{s_t})\}$ is the set of symptom names and associated status in a conversation, and $\alpha$ is the relative weight of the loss of the span extraction task and the attribute prediction task.

During training, we are simultaneously attempting to detect the location of tags as well as classify the tags. Initially, our model for locating the tags is unlikely to be reliable, so we adopt a curriculum learning paradigm. Specifically, we provide the classification stage the reference location of the tag from the training data with probability $p$, and the inferred location of the tag with probability $1-p$. We start the joint multi-task training by setting this probability to $1$ and decrease it as training progresses~\cite{BenVinJaiSha15}. 

\paragraph{Remarks} Although the SA-T model was developed to infer symptoms and status in the clinical domain, the formulation is general and can be applied to any domain. The model breaks up the task into identifying spans of interests and then classifying the span with richer contextual representations. The first stage alleviates data sparsity by pooling all spans of interest. When the label space naturally partitions into separate categories, the second stage can be broken up further into separate prediction tasks and reduces data splitting. 

\subsection{Sequence-to-sequence (Seq2Seq) Model}
\label{ssec:Seq2Seq}

As shown in Table \ref{table:multi_turn_examples}, symptoms are sometimes not stated explicitly, but rather explained or described in informal language over several conversation turns.  There may not even be a symptom entity that is explicitly mentioned; instead, the physician, as well as any symptom extraction model, must infer it from a description. To better capture symptoms that are not referred to by name, we explore an alternative formulation of the problem.  In this formulation, the input to the model is a chunk of the conversation, consisting of multiple consecutive turns from the doctor-patient conversation, and the output is a list of symptoms mentioned as well as their statuses.  The key difference between this formulation and the span-attribute tagging formulation is that the symptom entity is not assigned to a word or phrase in the input text. 

\begin{table}[t]
\centering
{\footnotesize
  \begin{tabular}{ l|l }
    Transcript & Symptoms + Status \\
    \hline
    \textbf{DR}: Any issues with your eyes? & \textbf{Eye pain}:  \\
    \textbf{PT}: Well sort of   & experienced   \\
    \textbf{DR}: Is your vision ok? & \textbf{Vision loss}:\\
    \textbf{PT}: Yeah, but the right one  hurts & not experienced \\
    \hline
    \textbf{DR}: How is your bladder? & \textbf{Frequent urination}: \\
    \textbf{PT}: I have to go, all the time & experienced \\
    \textbf{DR}: At night?  & \textbf{Nocturia}: \\
    \textbf{PT}: No, just during the day & not experienced \\
    \hline
  \end{tabular}
}
  \caption{Two illustrative examples where symptoms and their status are not described explicitly but need to be inferred from the context spanning multiple turns.}
\label{table:multi_turn_examples}
\vspace{-.2in}
\end{table}

In this formulation, each input example consists of a segment of transcript, represented as a sequence of tokens $\mb{x} = (x_1, ..., x_m)$, and a list of symptoms and their corresponding status $\mb{y} = (y_1, ..., y_n)$. Hence, it is well-suited to the sequence-to-sequence (Seq2Seq) class of models~\cite{Sutskever14, ChoMerrienboerGulcehreEtAl14} which has been successful across a variety of language understanding tasks, including conversation modeling \cite{Vinyals15}, abstractive summarization \cite{Nallapati16}, and question answering \cite{Seo17}. Following the standard Seq2Seq setup, our model is composed of two recurrent neural networks (RNNs), an encoder and a decoder. First, the \emph{encoder} consumes $\mb{x}$ one token at a time, producing an encoding, $\bsy{h}(x_i)$, for each token $x_i$.  Then the \emph{decoder} estimates an output distribution over sequences of symptoms and their status $\mb{y}$, conditional on the encodings.  An attention mechanism \cite{Bahdanau14} allows the decoder to combine information from the encoded sequences differently at each decoding step.

The Seq2Seq model is trained using a cross-entropy criterion to maximize $P(\mb{y}|\mb{x})$ -- the likelihood of reference symptoms and their status given the conversation transcripts.  At inference time, the most likely sequence of symptoms and their status is decoded one token a time using beam search. One challenge for Seq2Seq models is handling very long inputs \cite{Sutskever14}.  Therefore, unlike the span-attribute tagging model where each input example may be a full transcript, we use transcript segments consisting of $k$ consecutive turns. In practice we found a value of $k = 5$ to work well. A value of $k$ that is too small won't be enough to resolve symptoms like those in Table \ref{table:multi_turn_examples}, while a value of $k$ that is too large may degrade quality and make our model harder to train. At inference time, we use a sliding window of size $k$ across the full conversation, and then aggregate the predictions from those windows.

\subsection{Encoder Pre-training}
\label{ssec:encoders}

While the span-attribute tagging  and Seq2Seq models have different output layers, they use a common input encoder architecture. At any given input time, the conversation up to that time is represented by the hidden state of the encoder, which is used for making output predictions. We investigated two variations of the encoder. 

First, we compare the LSTM encoder with the Transformer encoder~\cite{transformer}. The key difference between them is that the LSTM relies on latent variables to propagate state information while Transformer relies solely on an attention mechanism. In a machine translation benchmark, the Transformer has been shown to outperform the LSTM encoder~\cite{transformer}, and a hybrid model, consisting of a Transformer encoder and an LSTM decoder, performed even better~\cite{mia}. We therefore compare the hybrid model, with the LSTM-only encoder-decoder model on our task. 

Second, we use a {\it pre-training} technique to leverage unlabeled data and improve the feature representation learned by the encoder \cite{Kannan18}. Given a short snippet of conversation, the model is tasked with predicting the next turn, similar to Skip Thought~\cite{Ryan_2015}. Since this task requires no labeling, the model can be trained on the full corpus of 90K conversations. The resulting encoder is plugged into our model for the symptom prediction task, and the full model is trained on the subset that is labeled. The pre-training can be performed for both the LSTM and Transformer encoders, as well as for both the Seq2Seq and the span-attribute tagging models. We did not experiment with alternative pre-training loss such as BERT~\cite{Devlin18}.

\section{Empirical Evaluations}
\label{sec:expts}

Before creating dedicated models for this task, we investigated general purpose named-entity annotation tools akin to \cite{Momchev:2010,Nothman:2008}. While a few of these tools can annotate symptom entities with some accuracy, they have no mechanism to infer the symptom status, which is required for clinical documentation. 

In all the experiments described below, our models were trained and evaluated on the corpus described in Section~\ref{ssec:corpus} using the metrics defined in Section~\ref{ssec:metrics}. Since our ontology differs from the public domain {\it i2b2} task, we could not evaluate our models on that task.

For a robust estimate of the model performance, the model outputs were evaluated against a ``voted'' reference created using the labels from three independent scribes. This is the case for all the results reported in the experiments below, unless otherwise specified. While our application requires jointly inferring both the symptom and status (Sx + Status), for a better understanding of the model behavior we have also included the performance on inferring just the symptom names (Sx). These are reported in separate columns in the tables below.

\subsection{Hyperparameters}
\label{ssec:optimalparams}

The hyperparameters of the Span-Attribute tagging (\SAT) and the Seq2Seq models were picked to maximize the performance on the development set. The models were trained using the Adam optimizer~\cite{KinBa15} and the selected parameters are reported in Table~\ref{table:ParamSeq2Seq}. 
\begin{table}[H]
\renewcommand\tabcolsep{4pt}
  \centering
  \begin{tabular}{llll}
  \hline
  Parameter & \SAT & Seq2Seq & Range \\ \hline
  Word emb & 256 & 256 & [128 -- 512] \\
  LSTM Cell & 1024 & 512 & [256 -- 1024] \\
  Enc/dec layers & 1 & 1 & [1 -- 3] \\
  Dropout & 0.4 & 0.0 & [0.0 -- 0.5] \\
  L2 & 1e-4 & 1e-4 & [1e-5 -- 1e-2]\\
  Std of VN & 1e-3 & 0.2 & [1e-4 -- 0.2] \\
  $\alpha$ of \SAT & 0.01 & n/a & [1e-4 -- 0.1]\\
  Learning rate & 1e-2 & 3e-3 & [1e-4 -- 1e-1]\\
  \hline
  \end{tabular}
  \caption{The range over which hyperparameters were tuned and the optimal choice for each model.}
  \label{table:ParamSeq2Seq}
\end{table}

\subsection{Different Encoders and Pre-training}

To select the encoder, first we evaluate the impact of pre-training on the LSTM encoder, using the Seq2Seq model. The results are reported in Table~\ref{table:pretraining}. The results show that pre-training of the LSTM encoder consistently improves performance of the Seq2Seq model across all metrics.

\begin{table}[ht]
  \centering
  \renewcommand\tabcolsep{4pt}
  \begin{tabular}{lcc}
  \hline
  Pretrained  & Sx & Sx + Status \\ \hline
  \multicolumn{3}{c}{\it Unweighted F1(Precision, Recall)} \\
  No  & {\bf 0.69} (0.66, 0.73) & {\bf 0.54} (0.49, 0.60) \\
  Yes & {\bf 0.70} (0.66, 0.75) & {\bf 0.55} (0.49, 0.62) \\
  \hline
  \hline
  \multicolumn{3}{c}{\it Weighted F1(Precision, Recall)} \\
  No  & {\bf 0.77} (0.76, 0.78) & {\bf 0.63} (0.60, 0.65) \\
  Yes & {\bf 0.79} (0.77, 0.80) & {\bf 0.64} (0.61, 0.68) \\
  \hline
  \end{tabular}
  \caption{The comparison of Seq2Seq model performance when the LSTM encoder is intialized randomly and when the encoder is pre-trained on the entire corpus including the unlabelled data.}
  \label{table:pretraining}
\end{table}

Next, the Transformer encoder was compared against the LSTM encoder, using pre-training in both cases. Based on the performance on the development set, the best encoder was chosen which consists of two layers, each with 1024 hidden dimension and 16 attention heads. 
The results in Table \ref{table:LSTMvsTransformer} show that the LSTM-encoder outperforms the Transformer-encoder consistently in this task, when both are pre-trained. Therefore, for the rest of the experiments, we only report results using the LSTM-encoder.

\begin{table}[t]
  \centering
  \begin{tabular}{lcc}
  \hline
  Encoder & Sx & Sx + Status \\ \hline
  \multicolumn{3}{c}{\it Unweighted F1(Precision, Recall)} \\
  Xformer & {\bf 0.67} (0.66, 0.67) & {\bf 0.51} (0.48, 0.54) \\
  LSTM & {\bf 0.70} (0.66, 0.75) & {\bf 0.55} (0.49, 0.62) \\
  \hline
  \hline
  \multicolumn{3}{c}{\it Weighted F1(Precision, Recall)} \\
  Xformer & {\bf 0.76} (0.79, 0.74) & {\bf 0.61} (0.62, 0.61) \\
  LSTM & {\bf 0.79} (0.77, 0.80) & {\bf 0.64} (0.61, 0.68) \\
  \hline
  \end{tabular}
  \caption{The comparison of Seq2Seq model performance using Transformer (Xformer) and LSTM encoders. Both encoders were pre-trained.}
  \label{table:LSTMvsTransformer}
\end{table}


\subsection{Manual Transcript Evaluation}
\label{ssec:multiplerefs}

Next, we evaluate and compare the performance of the models when they are trained and tested on the manual transcripts. For comparison, we include a standard tagging baseline consisting of a bidirectional LSTM-encoder (pre-trained as described in Section~\ref{ssec:encoders}), followed by two feed-forward layers and a softmax layer. The targets consisted of the cross product space of 186 symptom names and 3 status values. The model was trained using cross-entropy loss. Due to the large cross product label space, the CRF loss is infeasible in this setting. 


\begin{table}[ht]
  \centering
  \begin{tabular}{lcc}
  \hline
  Model   & Sx & Sx + Status \\ \hline
  \multicolumn{3}{c}{\it Unweighted F1(Precision, Recall)} \\
  Baseline & {\bf 0.68} (0.73, 0.63) & {\bf 0.50} (0.54, 0.47) \\
  \SAT & {\bf 0.71} (0.73, 0.69) & {\bf 0.58} (0.58, 0.58) \\
  Seq2Seq & {\bf 0.70} (0.66, 0.75) & {\bf 0.55} (0.49, 0.62) \\
  \hline
  \hline
  \multicolumn{3}{c}{\it Weighted F1(Precision, Recall)} \\
  Baseline & {\bf 0.73} (0.78, 0.69) & {\bf 0.57} (0.61, 0.53)\\
  \SAT & {\bf 0.77} (0.80, 0.74) & {\bf 0.65} (0.66, 0.63) \\  
  Seq2Seq & {\bf 0.79} (0.77, 0.80) & {\bf 0.64} (0.61, 0.68) \\
  \hline
  \end{tabular}
  \caption{The comparison of performance on manual transcripts between the baseline, the \SAT and the Seq2Seq models.}
  \label{table:ManualTxSingleRef}
\end{table}

From the results reported in Table~\ref{table:ManualTxSingleRef}, we see that the span-attribute tagging model performs as well as the Seq2Seq model. This is surprising since it is designed to not only predict the symptom name and status, but also to locate the words associated with them, a more demanding task. Another noteworthy difference between the two models is that the tagging model consistently trades off lower recall for higher precision, compared to the Seq2Seq model. The Mann-Whitney rank test indicates that improvements of both the models over the baseline are statistically significant under both metrics. In general, a gain of about 0.02 or more in F1-score was found to be statistically significant in our experiments on this task.

Knowing that the quality of the reference impacts the measured performance, we compared the model output to two versions of references in addition to the ``voted'' reference. In one version, we used a single reference for each conversation from a randomly chosen scribe. In another version, the model was given credit when the output matches ``any'' of the three scribes. This was motivated by the observation during adjudication that the symptom names may be annotated in more than one way, as illustrated in the example in Table~\ref{table:any}.

\begin{table}[h]
\centering
{\footnotesize
  \begin{tabular}{ l}
\textbf{PT:} I found the exercises very difficult. \\
\textbf{DR:} Was it \ul{hurting} you? \\
\textbf{PT:} Yeah, a lot. \\
  \end{tabular}
}
  \caption{An illustrative example to show how symptom ({\it hurting}) may be assigned either symptom names -- {\it sym:musculo\_skeletal:pain} or {\it sym:constitutional:pain}, which are both valid given the context.}
\label{table:any}
\end{table}

The model outputs were compared against the above mentioned variants of the reference and the results are reported in Table~\ref{table:SingleVsMultipleRefs}. The measured gap in performance between single reference and ``voted'' reference is small. The ``voted'' version corrects the reference, when one of the three scribes misses the annotation. However, when two scribes pick different valid labels and the third misses them, the ``voted'' reference is not better than the single reference. In such instances, allowing a model to match ``any'' of the references would be a reasonable solution. This may explain why the performance in that case is substantially better than the single or ``voted'' reference.
 
\begin{table}[t]
  \centering
  \renewcommand\tabcolsep{3pt}
  \begin{tabular}{crcc}
    \hline
    Model & Type & Sx & Sx + Status \\ \hline
    & \multicolumn{3}{c}{\it Unweighted F1(Precision, Recall)} \\
    \multirow{3}{*}{\rotatebox{90}{\SAT}} & Single & {\bf 0.70} (0.72, 0.69) & {\bf 0.56} (0.56, 0.57) \\
    & Voted  & {\bf 0.71} (0.73, 0.69) & {\bf 0.58} (0.58, 0.58) \\
    & Any  & {\bf 0.81} (0.84, 0.78) & {\bf 0.69} (0.71, 0.67)\\
    \hdashline
    \multirow{3}{*}{\rotatebox{90}{Seq2Seq}} & Single & {\bf 0.68} (0.62, 0.76) & {\bf 0.53} (0.45, 0.63) \\
    & Voted & {\bf 0.70} (0.66, 0.75) & {\bf 0.55} (0.49, 0.62) \\
    & Any & {\bf 0.81} (0.77, 0.84) & {\bf 0.67} (0.62, 0.73) \\
    \hline
    \hline
    & \multicolumn{3}{c}{\it Weighted F1(Precision, Recall)} \\
        \multirow{3}{*}{\rotatebox{90}{\SAT}} & Single & {\bf 0.76} (0.79, 0.74) & {\bf 0.63} (0.64, 0.62)\\
    & Voted  & {\bf 0.77} (0.80, 0.74) & {\bf 0.65} (0.66, 0.63)\\
    & Any  & {\bf 0.86} (0.89, 0.83) & {\bf 0.75} (0.77, 0.73)\\
        \hdashline
    \multirow{3}{*}{\rotatebox{90}{Seq2Seq}} & Single & {\bf 0.77} (0.73, 0.80) & {\bf 0.62} (0.57, 0.68)\\
    & Voted  & {\bf 0.79} (0.77, 0.80)  & {\bf 0.64} (0.61, 0.68)\\
    & Any  & {\bf 0.87} (0.86, 0.89) & {\bf 0.75} (0.72, 0.78) \\
    \hline
  \end{tabular}
  \caption{The comparison of model performance on manual transcripts when the performance was evaluated against \textit{Single}, \textit{Voted} and \textit{Any} reference labels.}
  \label{table:SingleVsMultipleRefs}
\end{table}


\subsection{ASR vs.~Manual Transcript Evaluation}
\label{ssec:asrtranscripts}
In clinical applications, manual transcripts will be unavailable and the model needs to infer symptom and status on transcripts obtained from an automatic speech recognition (ASR) system. We investigated the impact on performance when the test data is switched from manual to the corresponding ASR transcripts. Such a switch is expected to degrade the performance of models trained on manual transcripts and often this degradation can be alleviated by training the model  on ASR transcripts. So, we measured performance using models trained on different combinations of manual and ASR transcripts. 

Recall, the symptom, as described in Section~\ref{ssec:corpus}, were annotated on manual transcripts. These annotations were automatically transferred to the ASR transcripts by aligning the words in both transcripts for the same speaker turns and mapping the labels from manual transcripts to the corresponding words in the ASR transcripts. The word error rate of the ASR transcripts is about 20\%~\cite{CC_2018}. In the alignment process, a fraction of the labels (9.1\%) failed to alignment properly and were discarded. 

\begin{table}[t]
  \centering
  \renewcommand\tabcolsep{4pt}
  \begin{tabular}{crcc}
  \hline
  Model & Type & Manual Test& ASR Test\\ \hline
   & \multicolumn{3}{c}{\it Unweighted F1(Sx, Sx+Status)} \\
   \multirow{3}{*}{\rotatebox{90}{\SAT}} & Manual Train & { 0.71, 0.58} & {{\bf 0.67}, {\bf 0.52}} \\
&   ASR Train & { 0.68, 0.55} & {0.67, 0.52} \\
&  Combined & { {\bf 0.72}, {\bf 0.59}} & {0.66, 0.53} \\
    \hdashline
  \multirow{3}{*}{\rotatebox{90}{Seq2Seq}} & Manual Train & { 0.70, 0.55} & { 0.65, 0.50} \\
  & ASR Train & { 0.67, 0.50} & { 0.62, 0.47} \\
  & Combined & { 0.69, 0.54} & { 0.64, 0.49} \\
    \hline
    \hline
  &\multicolumn{3}{c}{\it Weighted F1(Sx, Sx+Status)} \\
  \multirow{3}{*}{\rotatebox{90}{\SAT}} & Manual Train & { 0.77, 0.65} & { 0.72, 0.58} \\
 & ASR Train & { 0.75, 0.62} & { 0.72, 0.58} \\
& Combined & { 0.78, 0.65} & { 0.71, 0.58} \\
    \hdashline
\multirow{3}{*}{\rotatebox{90}{Seq2Seq}} & Manual Train & {{\bf 0.79}, {\bf 0.64}} & { {\bf 0.75}, {\bf 0.59}} \\
& ASR Train & { 0.76, 0.61} & { 0.72, 0.57} \\
& Combined & { 0.79, 0.64} & { 0.74, 0.59} \\
    \hline
  \end{tabular}
  \caption{The comparison of model performances when trained on manual (Manual Train), ASR (ASR Train), and their combined (Combined) transcripts and evaluated on manual (Manual Test) and ASR (ASR Test) transcripts. The best performance is shown in bold.}
  \label{table:ManualVsASRTx}
\end{table}
The results, reported in Table~\ref{table:ManualVsASRTx} with ``voted'' reference, show that the performance of the models trained on manual transcripts ({\it Manual Train}) degraded when tested on ASR transcripts ({\it ASR Test}), for both models, as expected. But,  surprisingly, training models on ASR transcripts ({\it ASR Train}) or folding the ASR transcripts into the manual training data ({\it Combined}) did not improve the performance much. This maybe due to the fact that our performance metrics are evaluated at the conversation level and there is redundancy in clinical conversations, where the same symptom may be mentioned multiple times during the course of the conversation and each time in a different way.

\subsection{Symptom Names vs.~Body Systems}
\label{ssec:NameVsBodySystem}
One way to understand the confusion between symptom names is to measure the performance after projecting the inferred symptom names (186 types) to their corresponding body systems (14 types). For example, {\it sym:musculo-skeletal:pain} and {\it sym:musculo-skeletal:swelling} were collapsed to {\it sym:musculo-skeletal}.

As a baseline, we trained an LSTM tagger with a CRF output layer to predict targets consisting of the simple Cartesian product of symptom body systems and their status. The performance of the baseline system and our models were evaluated on manual transcripts. Our models were trained to predict the symptom name and the predictions were projected to the system level. The results are reported in  Table~\ref{table:NameVsSystem}.

\begin{table}[t]
  \centering
  \renewcommand\tabcolsep{4pt}
  \begin{tabular}{lcc}
  \hline
  Model & Sx + Status & Sx System + Status\\ 
  \hline
   \multicolumn{3}{c}{\it Unweighted F1(Precision, Recall)} \\
  Baseline & n/a & {\bf 0.60} (0.67, 0.54)\\ 
   \SAT & {\bf 0.58} (0.57, 0.58) & {\bf 0.69} (0.70, 0.69) \\
  Seq2Seq & {\bf 0.55} (0.49, 0.62) & {\bf 0.67} (0.62, 0.73) \\
  \hline
  \hline
  \multicolumn{3}{c}{\it Weighted F1(Precision, Recall)} \\
  Baseline & n/a & {\bf 0.68} (0.75, 0.62)\\
  \SAT & {\bf 0.65} (0.66, 0.63) & {\bf 0.77} (0.79, 0.76) \\
  Seq2Seq & {\bf 0.64} (0.61, 0.68) & {\bf 0.78} (0.76, 0.81) \\
  \hline
  \end{tabular}
  \caption{The comparison of model performances when the symptom names (Sx) are collapsed to their respective body system (Sx System) categories.}
  \label{table:NameVsSystem}
\end{table}

When the symptom names are collapsed into broader body systems, the performance improves as expected. The gain in performance is surprisingly large at about 0.14 F1-score. This suggests that a large fraction of confusion comes from names in the same body system. The baseline model has much lower precision and recall compared to our proposed models, even though it was trained on the body system labels directly, once again, demonstrating that the cross-product space is too sparse to be learned properly.

\section{Analysis}
\label{sec:analysis}
In this section, we conduct detailed comparisons among human scribes and our models.
\subsection{Human Performance}
\label{sssec:human}

To understand the inherent difficulty of this task, we estimated the human performance on this task by comparing each scribe against the reference generated from the ``voted'' results of the three scribes. Even though this estimate is inflated, because each scribes' annotation was counted towards the voted reference, it is a useful approximation. The results in Table~\ref{table:human} show two clear trends. First, even humans have difficulty identifying symptoms consistently. 
\begin{table}[t]
  \centering
  \begin{tabular}{lcc}
  \hline
  Model          & Sx & Sx + Status \\ \hline
  \multicolumn{3}{c}{\it Unweighted F1(Precision, Recall)} \\
  Human & {\bf 0.84} (0.86, 0.82) & {\bf 0.78} (0.80, 0.76)\\
  \SAT  & {\bf 0.71} (0.73, 0.69) & {\bf 0.58} (0.58, 0.57) \\
  Seq2Seq  & {\bf 0.70} (0.66, 0.75) & {\bf 0.55} (0.49, 0.62) \\
  \hline
  \hline
  \multicolumn{3}{c}{\it Weighted F1(Precision, Recall)} \\
  Human & {\bf 0.86} (0.88, 0.85) & {\bf 0.81} (0.82, 0.79)\\
  \SAT  & {\bf 0.77} (0.80, 0.74) & {\bf 0.65} (0.66, 0.63) \\
  Seq2Seq  & {\bf 0.79} (0.77, 0.80) & {\bf 0.64} (0.61, 0.68) \\
  \hline
  \end{tabular}
  \caption{The comparison of performance of models and single scribes against the ``voted'' reference.}
  \label{table:human}
\end{table}
For example, ``constitutional pain'' (non-specific) and ``musculo-skeletal pain'' were top confusions for our models as well as humans. Second, when status is considered, humans have less trouble inferring it from the context than our models, losing only 0.05 on F1 (weighted), while our models dropped about 0.14. Improving status classification remains one of our future work.

\subsection{Attention Weights}

Next, we inspected the Seq2Seq model's attention weights to see whether the evidence is scattered across words and turns in the dialog. Indeed, through manual inspection, we found this to be true qualitatively, as illustrated in Table \ref{table:attention_weights}. In this example, the symptom ``sym:const:difficulty sleeping'' is not mentioned directly but is implied from the evidence scattered in the context. Future work could use these weights to further investigate errors. 

\begin{table}[h]
\centering
{\footnotesize
  \begin{tabular}{ l}
\textbf{DR:} How is your \ul{sleep}? \\
\textbf{PT:} Well, I have been \ul{waking} up a lot. \\
\textbf{DR:} How often would you say? \\
\textbf{PT:} \ul{Several} times a \ul{night}. \\
\textbf{DR:} That is a lot of \ul{waking} up! \\
  \end{tabular}
}
  \caption{Example of attention from Seq2Seq model, where words with attention weight of 0.05 or higher are underlined.}
\label{table:attention_weights}
\end{table}

\subsection{Error Analysis}

Grouping false negatives by their symptom name, we observed that both models struggled with the symptoms -- pain, malaise, fatigue, difficult sleeping,  weight loss/gain, and frequent urination.  As illustrated in Table \ref{table:high_evidence_spread}, these symptoms were often communicated through back-and-forth with the doctor and therefore may have required combining evidence from multiple turns, making the inference more difficult.

\begin{table}[h]
\centering
{\footnotesize
  \begin{tabular}{ l}
    \hline
    \textbf{Muscoloskeletal pain} \\
    \textbf{DR:} Does it hurt when you go like this? \\
    \textbf{PT:} No, that shoulder is fine. \\
    \textbf{DR:} So this side hurts, but that side, if you \\
    reach, there's no pain? \\
    \textbf{PT:} Yeah, really only this one has been sore.  \\
    \hline
    \textbf{Weight loss/gain} \\
    \textbf{DR:} Okay. So when you took these, it went up? \\
    \textbf{PT:} Well it was high, then I lost a few pounds. \\
    Then just, it's been really stressful, I've slipped. \\
    \textbf{DR:} So it went back up? \\
    \textbf{PT:} Yeah, it's been up and down. \\
    \hline
  \end{tabular}
}
  \caption{Examples of evidence spreading across multiple turns.}
\label{table:high_evidence_spread}
\end{table}

\section{Conclusions}
This paper describes a novel information extraction task, that of extracting the symptoms mentioned in clinical conversations along with their status. We describe our corpus, the annotation paradigm, and tailored evaluation metrics. We proposed a novel span-attribute tagging (\SAT) model and a variant of sequence-to-sequence model to solve the problem. The \SAT model breaks up the task into identifying spans of interests and then classifying the span with richer contextual representations. The first stage alleviates data sparsity by pooling all spans of interest. When the label space naturally partitions into separate categories, the second stage can be broken up further into separate prediction tasks and reduces data splitting. Although the SA-T model was developed to infer symptoms and status in the clinical domain, the formulation is general and can be applied to any domain. As an alternative, our Seq2Seq model is designed to infer symptom labels when the evidence is scattered across multiple turns in a dialog and is not easily associated with a specific word span. The performance of our models is significantly better than baseline systems and range from an F-score of 0.5 to 0.8 depending on the condition. When the models are trained on manual transcripts and applied on ASR transcripts, the performance degrades considerably compared to applying them on manual transcripts. Training the model on ASR transcripts or on both ASR and manual transcripts does not help bridge the performance gap. Our analysis show that the \SAT model has higher precision while Seq2Seq model has higher recall, thus the two models compliment each other. We plan to investigate the impact of combining the two models.


\section*{Acknowledgments}
This work would not have been possible without the help of a number of colleagues, including Gang Li, Mingqiu Wang, Laurent El Shafey, Hagen Soltau, Patrick Nguyen, Nina Gonzales, Diana Jaunzeikare, Philip Chung, Ashley Robson Domin, Lauren Keyes, Alvin Rajkomar, Justin Stuart Paul, Katherine Chou, Chris Co, Claire Cui, and Kyle Scholz.

\bibliography{sx_medical_entities}
\bibliographystyle{acl_natbib}

%
%

\end{document}